\documentclass[letter]{article}
\pdfoutput=1
\usepackage{hyperref}
\hypersetup{
  pdfinfo={
    Title={title},
    Author={author},
  }
}
\usepackage{pdfpages}
\usepackage{amsmath}
\usepackage{float}
\usepackage{graphicx}
\usepackage{placeins}
\begin{document}
\title{Robust Pollen Imagery Classification with Generative Modeling and Mixup Training}
%
%
\author{Jaideep Murkute \\
\\ \href{jaideepm.111@gmail.com}{jaideepm.111@gmail.com}}
%
%


%
\maketitle              
\begin{abstract}
Deep learning based approaches have shown great
success in image classification tasks and can aid greatly towards the fast and reliable classification of pollen grain aerial imagery. However, often-times deep learning methods in the setting of natural images can suffer generalization problems and yields poor performance on unseen test distribution. In this work, we present and a robust deep learning framework that can generalize well for pollen grain aerobiological imagery classification.

We develop a convolutional neural network based pollen grain classification approach and combine some of the best practices in deep learning for better generalization. In addition to commonplace apparoches like data-augmentation and weight regularization we utilize implicit regularization methods like manifold mixup to allow learning of smoother decision boundaries. We also make use of proven state-of-the-art architectural choices like EfficientNet convolutional neural networks. Inspired by the success of generative
modeling with variational autoencoders, we train models with a  richer learning objective which can allow model to focus on the relevant parts of the image. Finally, we create an ensemble of neural networks, for robustness of the test set predictions. 
Based on our experiments, we show improved generalization
performance as measured with weighted F1-score with aforementioned approaches. The proposed approach earned a fourth-place in the final rankings in the ICPR-2020 Pollen Grain Classification Challenge; with a 0.972578 weighted F1 score, 0.950828 macro average F1 score and 0.972877 recognition accuracy.

\end{abstract}

\section{Introduction}
\label{section:intro}
The task of detection of pollen and the type of the pollen is one of the tasks in field of aero-biology. Ability to perform such tasks can help with the detection of allergens and other infectious diseases\cite{baseline_ref}. Deep learning methods have shows great success in achieving state of the art results on many computer vision  tasks including image recognition \cite{alexnet}, \cite{eff_net}, \cite{mask_rcnn} etc.. However, as is very often the case, such improvements are only possible with deep learning when we have large amount of labeled data. For pollen grain image classification problems, this bottleneck was addressed by the \cite{ref_proc1} creating a large scale annotated dataset with class label as well as segmentation masks. However, in order to build a reliable and robust system for pollen grain classification, we need to tackle other potential pitfalls when developing deep learning methods like over-fitting to the training data distribution, bias behavior from the class frequency distribution etc. which may prevent generalization. Unsupervised learning of disentangled features with generative modeling realized with variational autoencoders (VAEs) has been shown to aid with this problem and exhibits better generalization performance \cite{vae_gen_1}, \cite{vae_gen_2}. Mixup based training in input and latent space has also shown great improvements in the generalization performance in the recent years \cite{mixup}, \cite{manifold_mixup}. Careful architectural choices have also been shown to be critical in the deep learning performance for image recognition in \cite{eff_net}, giving rise to the state-of-the-art EfficientNet class of models. In this work, we employ these methods along with traditional approaches in the deep learning for better convergence.


\section{Proposed approach}
\label{section:approach}
The proposed approach makes use of unsupervised learning of the disentangled visual features with generative modeling modeled with the variational auto-encoder architecture. Such learning approaches have shown promise to improve generalization on the downstream tasks. We also couple the modify the learning objective with mixup. Other traditional tricks employed to improve the generalization such as data augmentation, weight regularization, ensembling etc. will be discussed in the section \ref{section:implementation}.

The input images are passed through the encoder network. Based on the outputs of the encoder, we infer one of the four the class labels for the current input sample. From output of the encoder, we also extract the latent code of the variational autoencoder framework.  Latent code sample is taken based on the prior distribution assumption for the generative process and is passed to the decoder network. The decoder aims to reconstruct the segmented patch consisting only of the pollen grain, which we hypothesize can help network to focus on the relevant areas of the image, ignoring the backgroud. In the subsections below, we disuss our variational autoencoder formualtion and the mixup training formulation. The detailed network architecture is shown in the Figure 1.  

We also note that based on our experiments and local validation set evaluation, we did not find it significantly useful to explicitly segment the pollen grains from the background. Possible reason can be that many of the recent CNN architectures are robust enough to handle some minor variations in the background. Such capabilities are  even more enhanced when model is trained with augmented data which includes data perturbations and dropout.

\subsection{Variational Autoencoder}
For unsupervised learning of the visual features, we model the generative process with  variational autoencoder(VAE) framework. We aim to the learn the latent generative factors $z$ from the training data distribution $x$ by learning the likelihood $p_{\theta}(x|z)$. The latent variable z is assumed to come from a certain prior distribution $p(z)$. This likelihood will be approximated by the encoder part of the model. Given the intractability of exact posterior and to allow back-propagation of error gradients, a distribution $q_{\phi}(z|x)$ is used to  approximate the true posterior $p(z|x)$ using the reparameterization trick \cite{auto_var_bayes}. The whole framework is parameterized by the neural network architecture mentioned in detail in the Experiements section.

The inference and the generative networks are stacked to form a variational autoencoder. The objective of such VAE is to maximize the following evidence variational lower bound(ELBO) with respect to the parameters $\theta$ and $\phi$.

%
%
\begin{equation}
\begin{aligned}
  \label{eq:VAEObj}
  L_{\text{unsup}}= \frac{1}{N} \sum_{n=1}^{N} {q_{\phi}(z|x_{n})} [\log {p_{\theta}(x_{n}|z)}]
  - KL[q_{\phi}(z|x_{n})||p(z)] \\
\end{aligned}
\end{equation}
where $p(z)$
defines the prior distribution of 
$y$ and
KL($||$) stands for the non-negative Kullback-Leibler divergence between the prior and the approximate posterior. 
In standard practice, 
the prior $p(Z)$ 
is often assumed to be 
an isotropic Gaussian $\mathcal{N}_(0, 1)$, 
and the posterior $q_{\phi}(z|x)$ distributions are parametrised as Gaussians with a diagonal covariance matrix.
The first term of the Eq. \ref{eq:VAEObj} can be appproaximated with the reconstruction error between the  reconstruction generated by the decoder and the target and we model it with mean squared error formulation.  

For the supervised learning from the class labels to estimate the  class probability of given an input image, we minimize the supervised objective in Eq. \ref{eq:clf_obj} with the negative log likelihood with the binary cross-entropy loss formulation. 
\begin{equation}
\begin{aligned}
  \label{eq:clf_obj}
  \mathcal{L_\mathrm{sup}} = - \frac{1}{N} \sum_{n=1}^{N} \ log({p_{\theta}(\hat{y}_{n}|x_{n})})
\end{aligned}
\end{equation}
; where $\hat{y}$ is the estimated output probability of the input data sample belonging to each of the four classes.

Thus the effective objective function we minimize is combination of the unsupervised loss from Eq. \ref{eq:VAEObj} and the supervised loss mentioned in \ref{eq:clf_obj}.
\begin{equation}
\begin{aligned}
  \label{eq:clf_obj}
  \mathcal{L_\mathrm{final}} = \mathcal{L_\mathrm{unsup}} + \mathcal{L_\mathrm{sup}}
\end{aligned}
\end{equation}

\subsection{Mixup Training}
To make the model more robust and inspired by \cite{mixup} and \cite{manifold_mixup}, we impose mixup based learning objective where instead of just learning true labels of the data samples independently, model is also asked to learn the interpolated space between the data sample my mixing the data samples and the target variables. Such learning objective is known to have an implicit regularization effect and give rise to the smoother decision boundary which can potentially generalize better.  

For each mini-batch of size $B$ data samples , we shuffle and combine the random pairs of data samples and target variables: $(x_{i}, x_{j})$ and corresponding label pair $(y_{i}, y_{j})$ and perform a linear mixup which can be given by below formulation in \ref{eq:mixup_data} and \ref{eq:mixup_label}:   



\begin{equation}
\begin{aligned}
\label{eq:mixup_data}
   mixup(x_i, x_j) = \lambda \cdot (x_i) + 
                            (1 - \lambda) \cdot (x_j) 
\end{aligned}
\end{equation}

\begin{equation}
\begin{aligned}
\label{eq:mixup_label}
   mixup(y_i, y_j) = \lambda \cdot (y_i) + 
                            (1 - \lambda) \cdot (y_j)
\end{aligned}
\end{equation}

; where $\lambda \sim Beta(\alpha, \alpha)$, $\alpha$ is the shape parameter of the Beta distribution and needs is tuned as a hyper-parameter. Higher value of $\alpha$ would yield higher regularization effect and vice-versa. Such mixup can be performed in the input space or in the any of the latent space of the neural network. During training, for each mini-batch we select a layer at random and apply mixup on its output embedding. Since latent representations are mixed, we also mix the reconstruction targets by the equal probabilities determined by the sampled value of $\lambda$.

\section{Dataset}
\label{section:dataset}
We use the labeled pollen imagery from the Pollen-13k dataset introduced in \cite{ref_proc1} and can be accessed from \cite{dataset_url}. The dataset consists of more than 13,000 labeled images including three categories of pollen grains and the negative case of the debris. The three pollen grain categories present in the dataset are Corylus avellana i.e. well-developed pollen grains, Corylus avellana i.e. anomalous pollen grains and Alnus i.e. well-developed pollen grains; while the negative case of debris consists of bubbles, dust and any non-pollen detected objects. Such large scale annotated dataset is the key that makes success of deep learning method presented possible.   
We utilize the whole dataset without filtering out any data samples. We split the labeled dataset into the 80\% train and the remaining 20\% as the validation set while maintaining the equal class distribution in the two sets. As mentioned in the section \ref{section:approach}, we only make use of the original images and the segmented patches, and do not use segmentation mask images explicitly.

\section{Implementation Details}
\label{section:implementation} 
\par We experimented with a variety of different configurations to make the modeling and the hyper-parameter choices. We evaluate these methods based on local validation score based on the local validation score. The strategy for creating validation split is mentioned in the section \ref{section:dataset}. 
For the encoder part of VAE, we use the state of the art Efficientnet\cite{eff_net} architecture which showed significant improvement for image recognition accuracy on the ImageNet dataset. EfficientNet architecture is proposed by carefully choosing the depth, width and the resolution parameters of the network architecture. We use the EfficientNet with the ImageNet pretrained weights \cite{ref_effnet_github}. We use a 5 layer deep decoder architecture regardless of the Encoder architecture and hyper-parameters being used. At the end of the EfficientNet encoder, we use 2 fully connected layers of output sizes 512 and 4 to classify input into one of the four classes. 

To aid the generalization, we design a data augmentation pipeline with following sequence of image transformations: \\
1. Resize image to (256, 256) with bi-linear interpolation. \\
2. Random rotations with angles ranging between -40 to +40 degrees. \\
3. Zoom in or out with up to 20\% scale. \\
4. Flip horizontally with 50\% probability. \\   
5. Flip vertically with 50\% probability. \\
6. Random crop of size (224, 224). \\
7. Normalize to the ImageNet mean and the standard deviation. \\  
During test time, we only apply the transformation steps 1, 6 and 7 mentioned above.

To determine the final predictions on the test set, we use an ensemble of 4 such models. Output probabilities are estimated from each of the four models and averaged with equal weight to form the final classification decision. Two of the models in the final ensemble have EfficientNet-B3 architecture as the encoder and the other two have EfficientNet-B4 architecture. Two of the models(one B3 and one B4) models are trained with mixup $\alpha$ = 0.5 while the other two are trained with mixup $\alpha$ = 1.0. EfficientNet architecture is composed of multiple convolutional 'blocks'. During training, for each mini-batch we randomly select a 'block' index after which mixup is applied. EfficientNet B3 has 33 such blocks while B4 has 27 blocks, one of which is selected at random. Standard EfficientNet B3 and B4 architectures have dropout\cite{dropout} layer by with drop probability of 0.3 and 0.4 respectively. We retain this dropout layer which can act as an additional regularizer.  

We train all the models with mini-batch size of 64 samples, with the SGD optimizer with nesterov, momemtum of 0.9 and the L2 weight penalty  parameter of 1e-4. We use  the initial learning rate of 0.01 and is decayed with the factor of 0.1 if the loss plateaus for 10 epochs. For the first 30 epochs, we freeze the layers of the encoder part and just train the decoder and the classification branch based on the features extracted by the ImageNet pretrained encoder. Then we unfreeze the entire network and jointly trained further for 50 epochs.   

\section{Results}
\label{section:results}
\begin{figure}[H]
\centering
\parbox{5cm}{
\includegraphics[width=6cm]{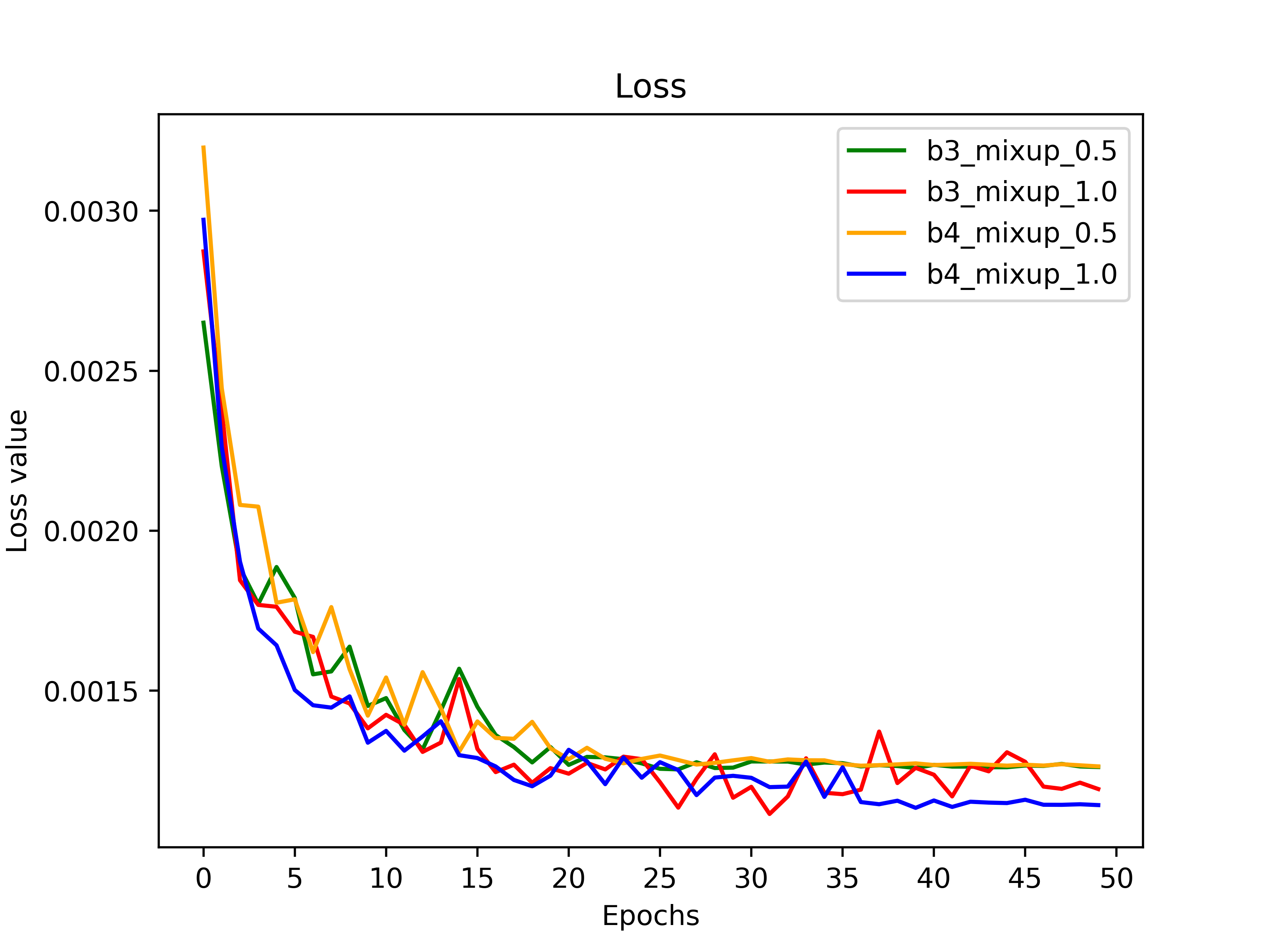}
}
\qquad
\begin{minipage}{5cm}
\includegraphics[width=6cm]{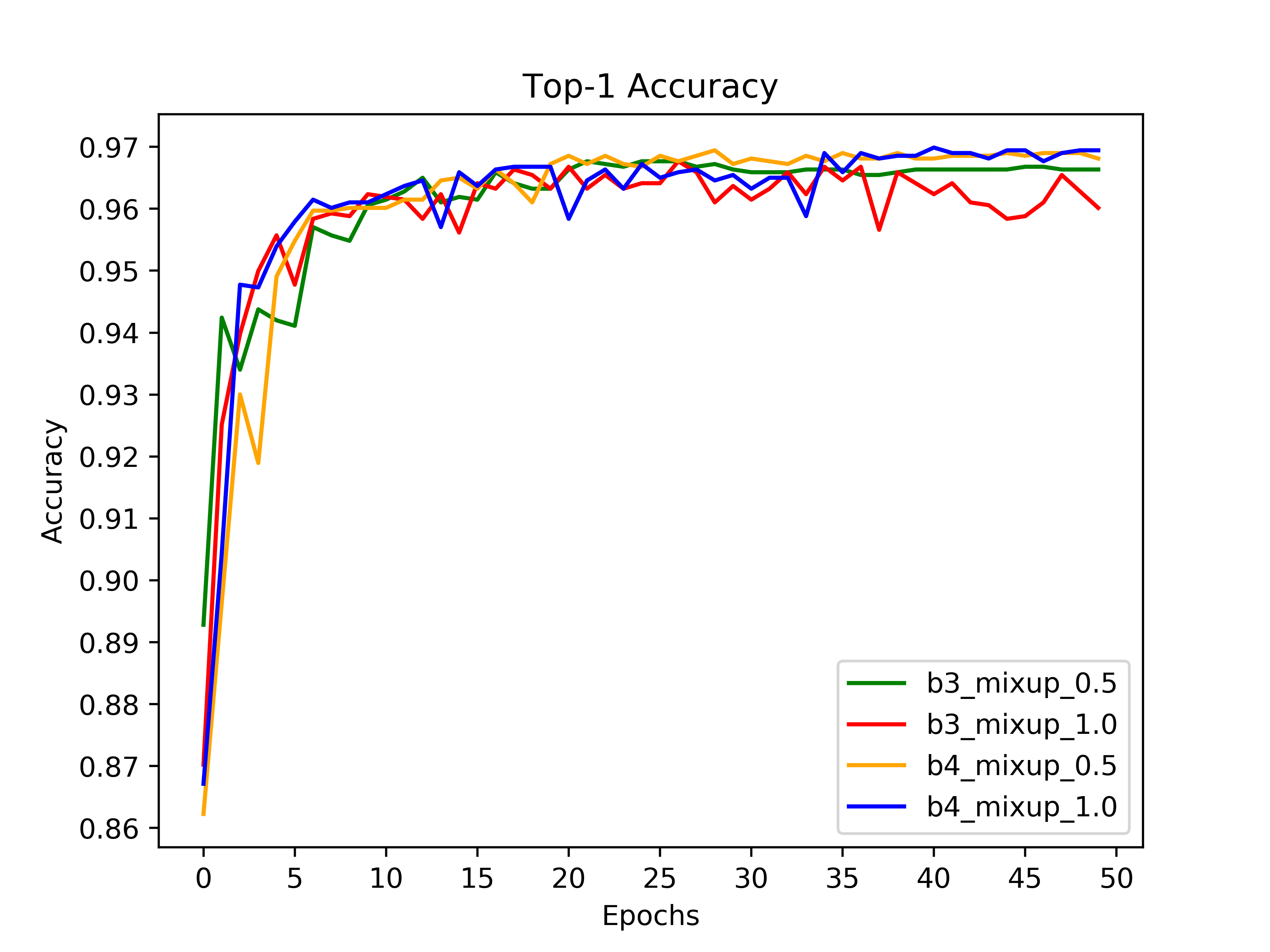}
\end{minipage}
\caption{[Best viewed in color at 200\% zoom] \textit{Left}: The aggregate loss value during the second stage when VAE model is trained with supervised and unsupervised losses together. \textit{Right}: Corresponding Top-1 accuracy values during training.}
\label{fig:loss_acc}
\end{figure}

Figure \ref{fig:loss_acc} shows the loss the accuracy values monitored during the training process. The model weights at the best epoch based on the validation accuracy value during training are stored and used for the final evaluation of the model.   
Results on the local validation set are shown the Table \ref{tab1}. The 'baseline' models are the EfficientNet B3 and B4 models without generative modeling approach i.e. simple CNN classifier without the decoder and also without the mixup. Since each of these models are tuned with the different random 20\% validation split to incorporate diversity in the models, we do not present the ensemble performance on the validation set. The ensemble formed by averaging outputs of the four VAE models in the Table \ref{tab1}, yields weighted  F1-score of 0.97258 and accuracy of 0.97287.

\setlength{\tabcolsep}{18pt}
\renewcommand{\arraystretch}{1.5}
\FloatBarrier
\begin{table}[h]
\label{tab: results_1}
\centering
\caption{Weighted-F1 scores of the four models in the final ensemble evaluated on the local validation set. 'Efficient Net' abbreviate as 'EffNet' for readability.}\label{tab1}
\begin{tabular}{|l|l|l|}
\hline
\textbf{Model} & \textbf{Accuracy} & \textbf{Weighted-F1 score} \\
\hline
EffNet-B3 (\textit{Baseline})    & 0.95970    & 0.95911 \\
EffNet-B4 (\textit{Baseline})    & 0.96147     & 0.96104 \\
EffNet-B3 VAE (mixup $\alpha$=0.5) & 0.96767    & 0.96731 \\
EffNet-B4 VAE (mixup $\alpha$=0.5) & 0.96988    & 0.96909 \\
EffNet-B3 VAE (mixup $\alpha$=1.0) & 0.96944    & 0.96953 \\
EffNet-B4 VAE (mixup $\alpha$=1.0) & 0.96767    & 0.96738 \\
\hline
\end{tabular}
\end{table}



\section{Conclusion}
\label{section:conclusion}
In this paper, we presented a robust deep learning based appraoch to classify pollen grain aerobiological images. The method combines the some of the well known approaches that are known to aid the generalization such as generative modeling with VAEs, mixup training, data augmentation, dropout and weight regularization etc. The method exhibits high  accuracy and the weighted F-1 score. The approach can potentially be improved further by performing in depth analysis of the failure modes on the edge cases and including segmentation masks in the model training process.

%
%
%
%

\end{document}